\documentclass[aps,prl,reprint,superscriptaddress,floatfix
]{revtex4-1}
\usepackage{graphicx}
\usepackage{color}
\usepackage{dcolumn}
\usepackage{bm}
\usepackage{amssymb}
\usepackage{color,amsmath}
\usepackage{soul,xcolor} 
\setstcolor{red}
\usepackage{siunitx}
\usepackage{epstopdf}
\usepackage{gensymb}
\usepackage{enumitem}

\DeclareSIUnit\angstrom{\text {Å}}

\begin{document}

\title{Zero-shot Medical Event Prediction Using a Generative Pre-trained Transformer on Electronic Health Records}


\author{Ekaterina Redekop}
\affiliation{Biomedical AI Research Lab, University of California, Los Angeles, Los Angeles, USA}
\affiliation{Department of Radiology, University of California, Los Angeles, Los Angeles, USA}
\affiliation{Department of Bioengineering, University of California, Los Angeles,  Los Angeles, USA}

\author{Zichen Wang}
\affiliation{Biomedical AI Research Lab, University of California, Los Angeles, Los Angeles, USA}
\affiliation{Department of Radiology, University of California, Los Angeles, Los Angeles, USA}
\affiliation{Department of Bioengineering, University of California, Los Angeles,  Los Angeles, USA}

\author{Rushikesh Kulkarni}
\affiliation{Biomedical AI Research Lab, University of California, Los Angeles, Los Angeles, USA}
\affiliation{Department of Radiology, University of California, Los Angeles, Los Angeles, USA}

\author{Mara Pleasure}
\affiliation{Biomedical AI Research Lab, University of California, Los Angeles, Los Angeles, USA}
\affiliation{Department of Radiology, University of California, Los Angeles, Los Angeles, USA}
\affiliation{Bioinformatics, University of California, Los Angeles, Los Angeles, USA}

\author{Aaron Chin}
\affiliation{Department of Pediatrics, Division of Immunology, Allergy and Rheumatology, University of California, Los Angeles, Los Angeles, USA}
\affiliation{Health Department of Information Services and Solutions, University of California, Los Angeles, University of California, Los Angeles, Los Angeles, USA}

\author{Hamid Reza Hassanzadeh}
\affiliation{Optum AI}

\author{Brian L. Hill}
\affiliation{Optum AI}

\author{Melika Emami}
\affiliation{Optum AI}

\author{William Speier}
\affiliation{Biomedical AI Research Lab, University of California, Los Angeles, Los Angeles, USA}
\affiliation{Department of Radiology, University of California, Los Angeles, Los Angeles, USA}

\author{Corey W. Arnold}
\email{CWArnold@mednet.ucla.edu}
\affiliation{Biomedical AI Research Lab, University of California, Los Angeles, Los Angeles, USA}
\affiliation{Department of Radiology, University of California, Los Angeles, Los Angeles, USA}
\affiliation{Department of Bioengineering, University of California, Los Angeles,  Los Angeles, USA}
\affiliation{Department of Computational Medicine, University of California Los Angeles, Los Angeles, CA, USA}

\maketitle
\textbf{Objectives:}
Longitudinal data in electronic health records (EHRs) represent an individual`s clinical history through a sequence of codified concepts, including diagnoses, procedures, medications, and laboratory tests. Generative pre-trained transformers (GPT) can leverage this data to predict future events. While fine-tuning of these models can enhance task-specific performance, it becomes costly when applied to many clinical prediction tasks. In contrast, a pretrained foundation model can be used in zero-shot forecasting setting, offering a scalable alternative to fine-tuning separate models for each outcome.

\textbf{Materials and methods:}
This study presents the first comprehensive analysis of zero-shot forecasting with GPT-based foundational models in EHRs, introducing a novel pipeline that formulates medical concept prediction as a generative modeling task. Unlike supervised approaches requiring extensive labeled data, our method enables the model to forecast a next medical event purely from a pretraining knowledge. We evaluate performance across multiple time horizons and clinical categories, demonstrating model`s ability to capture latent temporal dependencies and complex patient trajectories without task supervision.

\textbf{Results:}
Model performance for predicting the next medical concept was evaluated using precision and recall metrics, achieving an average top1 precision of 0.614 and recall of 0.524. For 12 major diagnostic conditions, the model demonstrated strong zero-shot performance, achieving high true positive rates while maintaining low false positives.

\textbf{Discussion and conclusion:}
We demonstrate the power of a foundational EHR GPT model in capturing diverse phenotypes and enabling robust, zero-shot forecasting of clinical outcomes. This capability enhances the versatility of predictive healthcare models and reduces the need for task-specific training, enabling more scalable applications in clinical settings.


\begin{figure*}[t]
\includegraphics  {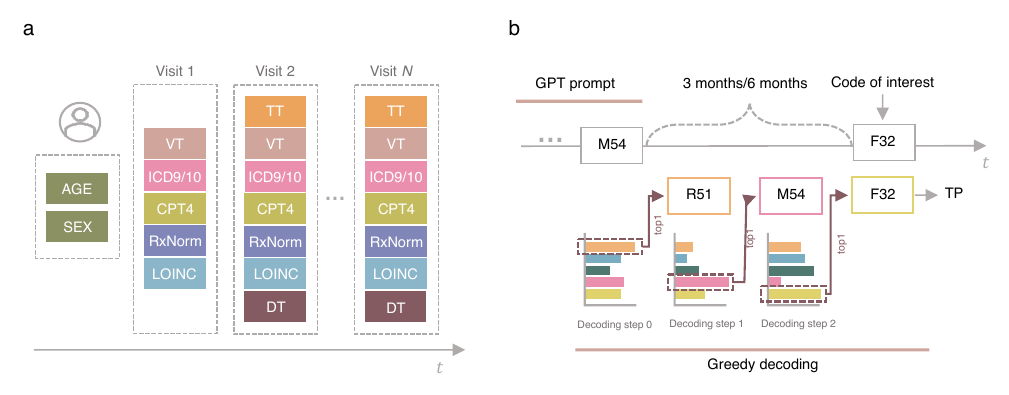}
\caption{\textbf{a} Representation of individual's EHR as a sequence of visits with demographics data including age and sex. Each visit can be represented as a sequence of concepts such as diagnoses (ICD9/10), procedures (CPT4), medications (RxNorm),  and labs (LOINC). Each visit, except the first one, starts with a relative time token (TT) followed by a visit type (VT). A visit can end with a discharge type token (DT) for inpatient visits. Each visit sequence is separated by a separation token. \textbf{b} An example of an individual's EHR timeline leading up to the appearance of a specific diagnostic code of interest (e.g., F32 - major depressive disorder). The timeline shows the individual's medical history ending at a point that is either 3 or 6 months before the actual occurrence of the target code. This period between the end of the shown history and the appearance of the target code represents the prediction window, during which the model attempts to forecast the likelihood of the target condition developing. The top 1 greedy decoding path was considered at each time step. }
\label{fig1}
\end{figure*}

\section*{\textbf{Introduction}} 
Electronic health records (EHRs) provide a longitudinal representation of an individual's current health status and clinical history using sequences of codes for various medical concepts, such as diagnoses, procedures, medications, and lab tests. Temporal modeling of EHRs can be used to predict future conditions and has been an active area of research with the goal of improving personalized medicine. Deep learning models based on transformer architectures have successfully predicted diseases and outcomes using EHRs for various applications like population estimation, disease progression analysis, and counterfactual reasoning \citep{yang2023transformehr,kraljevic2024foresight}. Existing approaches that leverage transformer architectures for EHR data can be broadly grouped into two categories: models utilizing the train-fine-tune paradigm \citep{yang2023transformehr,pang2024cehr,meng2021bidirectional} and models focused on forecasting \citep{kraljevic2024foresight}. The train-fine-tune paradigm involves pretraining a model on a large dataset, followed by supervised fine-tuning on a smaller, task-specific dataset to adapt the model for specific applications. Forecasting performs sequential prediction of the following most probable event given an individual`s history. 


TransformEHR \citep{yang2023transformehr} is a transformer-based encoder-decoder framework pretrained on 6.5 million EHRs to forecast future ICD codes. It was fine-tuned to predict 10 common and 10 uncommon diseases, including chronic PTSD, type 2 diabetes, and hyperlipidemia, outperforming encoder-only BERT models \cite{rasmy2021med}. However, the model is limited to ICD codes, excluding other critical codes for the prediction of outcomes. Additionally, it cannot predict event timing, since the time token is treated as an input embedding. The framework's reliance on fine-tuning necessitates separate models for each task, complicating deployment in practice.

Another recent work, Foresight \citep{kraljevic2024foresight}, introduced an EHR forecasting framework using the GPT-2 model \citep{radford2019language}. The model was trained and tested on three private datasets of 1.5 million, 27,929, and 46,520 individuals. Unlike previous work, it used biomedical concepts extracted from free text and linked to SNOMED, including diseases, symptoms, medications, findings, and other entities. The model's forecasting performance was evaluated using precision and recall for top-1 predictions, achieving 0.55 precision and 0.47 recall on the largest dataset. Considering the top 10 candidates, the precision improved to 0.84 and the recall to 0.76. However, the model's zero-shot capabilities for disease prediction were not evaluated, a large limitation of their approach.

Recent work, CEHR-BERT \citep{pang2024cehr}, utilized a GPT-type model to generate synthetic EHR data. The model was pretrained on 3.7 million individuals’ medical histories, including conditions, medications, procedures, visit types, discharge facilities, and temporal data like start year, age, visit intervals, and durations. The GPT model was trained to learn the distribution of EHR sequences to generate new synthetic sequences. Predictive performance on synthetic data was evaluated using logistic regression on five tasks, such as HF readmission and hospitalization, showing that synthetic data performed as well or better than real data. Similarly to the previous works, the model`s zero-shot capabilities were never evaluated.

Task-specific fine-tuning of pre-trained models for a wide range of conditions and diseases becomes impractical due to high computational costs, challenges in cohort labeling, and the continuous need to update models with new clinical data. Therefore, a general model that could approach the performance of many fine-tuned models would simplify development, lower costs, and increase implementation, ultimately resulting in a greater impact on patient care. Zero-shot evaluation is defined as the ability of a model to make accurate predictions on unseen data without requiring additional training, thus allowing it to screen for multiple conditions and forecast a range of outcomes simultaneously. We present a zero-shot evaluation of a GPT model trained to forecast EHR concepts and examine the prospective performance of the model on various predictive tasks, including 12 specific diagnostic conditions and 14 broader categories of medical conditions. While our focus is on zero-shot inference, we also report results from task-specific fine-tuning to benchmark potential gains from full supervision.

\section*{\textbf{Methods}} 

\subsection*{\textbf{Data}} 
Our cohort comprises two decades of history from 2.4 million individuals who received care at UCLA Health sites across the greater Los Angeles region (see TABLE \ref{tab:demographics}). To simulate a prospective deployment where the model would perform inference on data generated post-training, the model was trained using data before 2022 and tested with data acquired during and after 2022.  The pre-2022 data set was randomly split into a training set (90\%) and a validation set (10\%) at the patient level. Data is structured according to the Observational Medical Outcomes Partnership (OMOP) Common Data Model (CDM). The OMOP CDM standardizes the format and content of EHR data, facilitating data integration, sharing, and analysis across diverse healthcare systems.

\textcolor{red}{\begin{table}[ht]
\centering
\caption{Demographic characteristics of the study cohort.}
\label{tab:demographics}
\begin{tabular}{|p{5cm}|c|c|}
\hline
\textbf{Characteristic} & \textbf{Value} & \textbf{N (\%)} \\
\hline
\textbf{Age (years)} & Mean (SD) & 53.3 (20.0) \\
\textbf{Sex} & & \\
\quad Male & 820 & 49.2\% \\
\quad Female & 847 & 50.8\% \\
\textbf{Race} & & \\
\quad White & 1001400 & 44.9\% \\
\quad Black or African American & 99894 & 4.7\% \\
\quad Asian & 180447 & 8.1\% \\
\quad Native Hawaiian or Other Pacific Islander & 4665 & 0.3\% \\
\quad Multirace & 85774 & 3.8\% \\
\quad Other / Unknown & 849267 & 38.2\% \\
\hline
\end{tabular}
\end{table}
}
Diagnostic codes are encoded using the International Classification of Diseases, 9th Revision (ICD-9) and the International Classification of Diseases, 10th Revision (ICD-10). Procedure codes are represented using Current Procedural Terminology, 4th Edition (CPT-4). Medications are represented using RxNorm codes, and laboratory tests are recorded using Logical Observation Identifiers Names and Codes (LOINC). 

Data standardization involved converting ICD-9 codes to ICD-10 using General Equivalence Mappings, thereby enhancing consistency across individual`s records.
Full ICD-10 codes are high dimensional and are typically sparsely represented in individual`s sequences. Following previous works, dimensionality was reduced by grouping codes by the three digits before the decimal point \citep{choi2016doctor, meng2021bidirectional}.

Additionally, codes that appeared in less than 0.1\% of the database were removed from the dataset. Laboratory values were discretized using methods presented in the work by Bellamy et al. \citep{bellamy2023labrador}. After infrequent codes are removed, the empirical cumulative distribution function (eCDF) is computed using the training split for each unique lab code. We construct an integer-only token vocabulary by assigning a unique integer to each decile of the eCDF for each lab code.

The final vocabulary includes 1,204 unique diagnosis codes, 1,791 medication codes, 906 procedure codes, and 4,464 laboratory codes.

We used a context window of 512 tokens, truncating longer histories to fit the defined window.

The dataset used in this work was de-identified, and the study was therefore certified as exempt from IRB review by the UCLA IRB (IRB\#24-001116).

\subsection{Generative pre-trained transformer}

For each individual $u_{i}$ the full EHR sequence is organized as:
\begin{equation}
u_{i} = (V_1, \texttt{[SEP]}, V_2, \texttt{[SEP]}, \ldots, V_L),
\end{equation}
where each visit $V_t = (w_{t,1}, w_{t,2}, \ldots, w_{t,m_t})$ consists of $m_t$ medical events, demographics, and special tokens that characterize the visit (see FIG. \ref{fig1} a). Medical events include diagnostic $d$, medication $m$, procedure $p$, and laboratory codes $l$. The individual's overall timeline begins with demographic information, such as gender and age. Each visit starts with a special time token that indicates the time difference relative to the previous visit. This time token, denoted as $t$, can take one of the following values to indicate when the visit occurred relative to the previous visit: $t_{0}$, visit occurred within the same quarter; $t_{1}$, visit occurred more than three months but within six months; $t_{2}$, visit occurred later than six months but within a year; and $t_{3}$, visit occurred more than a year later. The time token is followed by a visit-type token $vt$ at the beginning of each visit. Each visit can end with discharge type tokens $dt$. Each event token $w_{v,j}$ is associated with a timestamp $\tau_{v,j}$ such that $\tau_{v,j} \leq \tau_{v,j+1}$ within the same visit. The objective of EHR modeling is then similar to language modeling:
\begin{equation}
\mathcal{L}(u_i) = \sum_{t=1}^{L} \sum_{j=1}^{m_t} \log P(w_{t,j} \mid w_{1,1}, \ldots, w_{t,j-1})
\end{equation}

\subsection*{\textbf{Evaluation cohort}} 
\label{sec:zero_shot_cohort}
We identified 12 specific diagnostic conditions to evaluate the zero-shot capabilities of the pretrained GPT model. These conditions include depression (F32, F33, F34), heart failure (I50), end-stage renal disease (N18), chronic ischemic heart disease (I25), type II diabetes (E11), chronic obstructive pulmonary disease (J44), pancreatic cancer (C25), liver cancer (C22), brain cancer (C71), cerebral infarction (I63), rheumatoid arthritis (M06), systemic lupus erythematosus (M32).

We then extended our evaluation to higher-level categories of codes to assess the model's ability to generalize across broader categories of medical conditions. Specifically, we included the following 14 categories: 'certain infectious and parasitic diseases' (A00-B99), 'neoplasms' (C00-D49), 'diseases of the blood and blood-forming organs and certain disorders involving the immune mechanism' (D50-D89), 'endocrine, nutritional and metabolic diseases' (E00-E89), 'mental, behavioral and neurodevelopmental disorders' (F01-F99), 'diseases of the nervous system' (G00-G99), 'diseases of the eye and adnexa' (H00-H59), 'diseases of the ear and mastoid process' (H60-H95), 'diseases of the circulatory system' (I00-I99), 'diseases of the respiratory system' (J00-J99), 'diseases of the digestive system' (K00-K95), 'diseases of the skin and subcutaneous tissue (L00-L99), 'diseases of the musculoskeletal system and connective tissue' (M00-M99), and 'diseases of the genitourinary system' (N00-N99).

\section*{\textbf{Metrics}} 
The model’s forecasting performance is evaluated at the visit level. For a given patient sequence, the first visit is used as input to generate the second visit, and the predicted visit is then compared to the actual second visit. Next, the first two observed visits are provided as input to generate the third visit, and so on. This stepwise evaluation continues through the entire timeline, ensuring that each prediction is made using only prior information.
At each step, the model autoregressively generates a sequence of medical concepts—such as diagnoses, procedures, medications, and labs—until a separation token is produced, indicating the end of the visit. We use greedy decoding (top-1 selection) at each time step to produce the output tokens. Precision and recall are computed by comparing the set of predicted tokens for the visit against the corresponding ground truth tokens, where precision is defined as $\frac{TP}{TP + FP}$ and recall as $\frac{TP}{TP + FN}$.

Additionally, we constructed 12 prediction tasks utilizing specific diagnostic conditions to evaluate the model's zero-shot capabilities. At each prediction window (three months and six months), the percentage of TP, FP, TN, and FN predictions are calculated (see FIG. \ref{fig1} b). Since the model can be a risk assessment tool, $N$ forecasts were considered at each step to estimate the amount of TP, FP, TN, and FN predictions (TP@$N$, FP@$N$, TN@$N$, FN@$N$) with $N=1,5,10,20$. 

Similarly, zero-shot performance was evaluated on the 14 higher-level diagnostic codes sections.


\subsection*{\textbf{Fine-tuning Setup for Diagnostic Prediction}}

To compare zero-shot forecasting with supervised learning, we conducted a fine-tuning experiment using a binary classification task for each evaluation cohort. We used the pretrained GPT model as a frozen feature extractor. The model’s output embeddings were pooled using mean pooling across tokens and passed through a dropout layer, followed by a linear classification head trained for binary prediction. Only the classification head was updated during training, while the transformer weights remained frozen. The model was trained using the Adam optimizer with a learning rate of 1e-4 with class balancing via weighted sampling.
We measured the number of true positives (TP) achieved at the same number of false positives (FP) as observed in the results of zero-shot evaluation for the same disease code. This allowed direct comparison of classification utility between fine-tuning and zero-shot evaluation under equal specificity constraints.

\section*{\textbf{Results}} 

\subsection*{\textbf{Zero-shot evaluation results}}

The average precision (at top 1) for forecasting all tokens was 0.614, and recall was 0.524.

\begin{figure*}[ht!]
\includegraphics{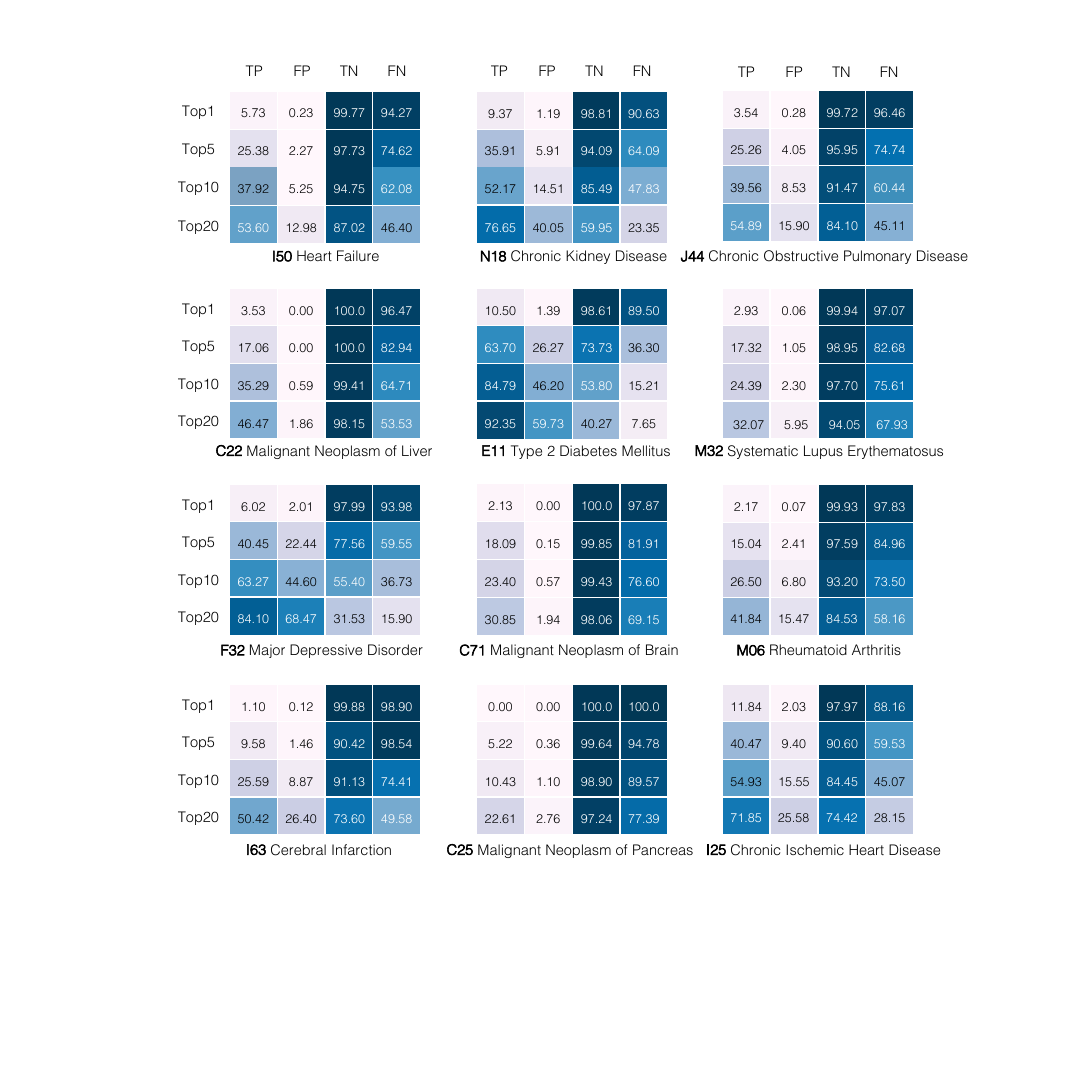}
\caption{\textbf{Zero-shot evaluation results of the GPT model for 12 specific diagnostic conditions at a prediction window of three months. Results are shown using TP,FP,TN and FN in \% for $N=1,5,10,20$.}
}
\label{fig3}
\end{figure*}

\begin{figure*}[ht!]
\includegraphics{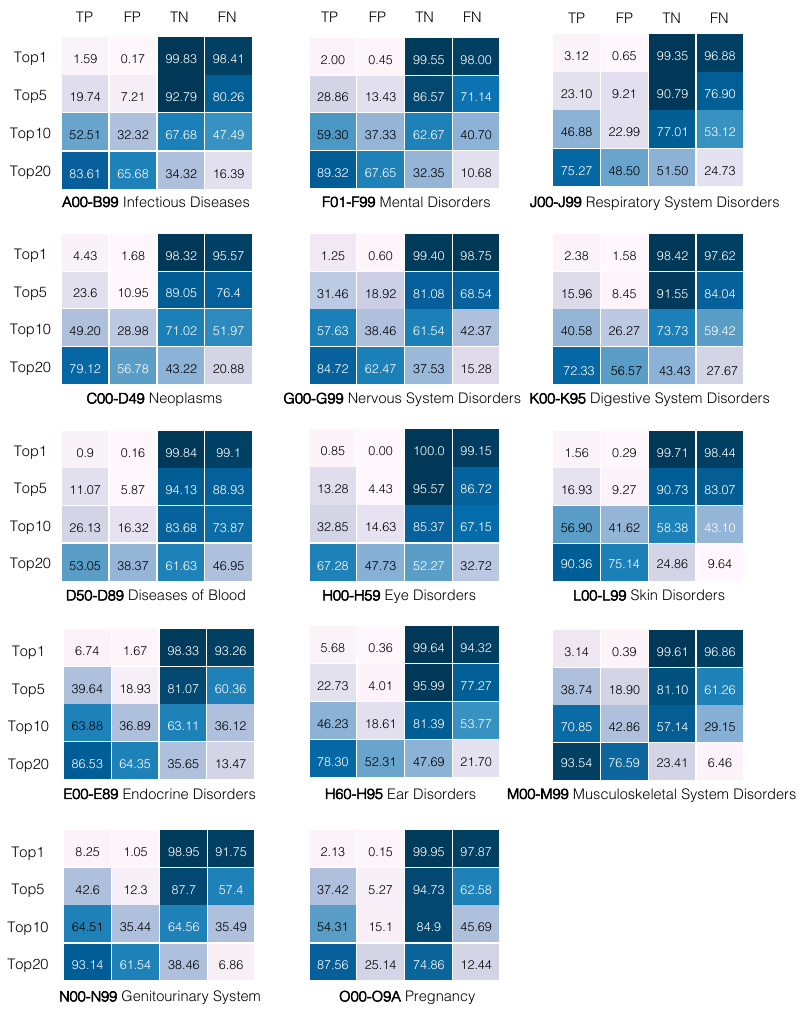}
\caption{\textbf{Zero-shot evaluation results of the GPT model for 14 higher-level categories at a prediction window of three months. Results are shown using TP,FP,TN and FN in \% for $N=1,5,10,20$.}
}
\label{fig5}
\end{figure*}

The evaluation results of the model's zero-shot capabilities for 12 diagnostic tasks at each prediction window (three months and six months) are presented in FIG.\ref{fig3},\ref{fig4} and for 14 higher-level diagnostic tasks in FIG.\ref{fig5},\ref{fig6}. The results demonstrate the model`s capability to predict multiple disease outcomes with varying levels of accuracy depending on the disease, prediction window, and the number of forecasts $N$ (e.g., top 1, top 5, top 10, or top 20).
Our analysis reveals that the model's predictive performance varies across different diseases and time horizons. As we expand the number of top predictions considered (from top 1 to top 20), we observe a consistent increase in TP across all conditions as it allows the model to evaluate a wider range of possible outcomes at each time step. However, a similar trend is observed for false positives, which also tend to rise as $N$ increases.
For example, the model predicted that 9.37\% of individuals diagnosed with chronic kidney disease would develop the condition within the three-month prediction window for $N$=1. When $N$ was increased to 20, the percentage of TP rose to 84.1\%, accompanied by a corresponding rise in FP to 68.47\% within the same three-month window. The same trend was observed for individuals diagnosed with chronic obstructive pulmonary disease. While 3.54\% of individuals diagnosed with the condition were predicted to be positive by the model for $N$=1, the percentage increased to 54.89\% as $N$ increased to 20. 
The model could predict 4.43\% TP individuals with diseases from the 'C00-D49 neoplasms' category at $N$=1, with this value increasing to 79.12\% as $N$ increased to 20. A corresponding increase was observed in the number of FP, rising from 1.68\% at $N$=1 to 56.78\% at $N$=20. 
Generally, the 3-month predictions show higher TP and less FP than the 6-month predictions across all conditions. An example that highlights this trend is the model's performance on heart failure, predicting that 5.73\% of individuals would experience heart failure within the three-month prediction window and 4.57\% within the six-month window for $N$ equal to 1. 
Only 0.16\% of individuals were FP within the three-month prediction window, while this figure increased slightly to 0.23\% for the six-month window. When $N=20$, forecasts at each time step lead to an increase in the number of TP individuals up to 53.6\% with a corresponding increase in FP up to 10.29\% within the three-month prediction window. The number of TP dropped slightly to 47.97\% for the six-month window, with a slight increase in FP to 12.98\%.



\begin{figure*}[ht!]
\includegraphics[width=0.95\linewidth]{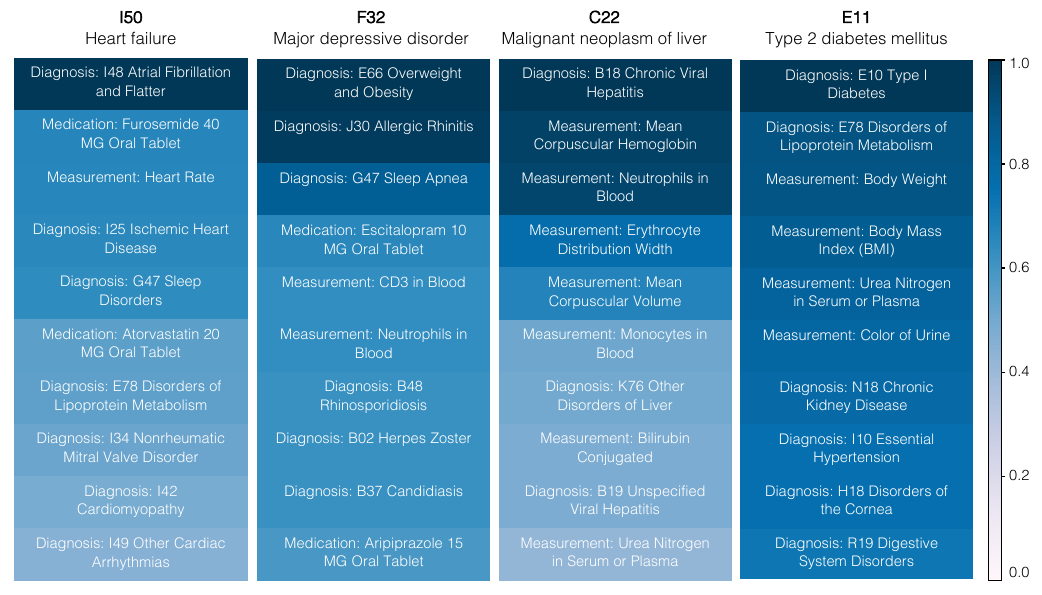}
\caption{\textbf{The 10 most frequent codes with the highest attention scores for 4 out of 12 specific diagnostic conditions, including heart failure, major depressive disorder, malignant neoplasm of liver, and type 2 diabetes mellitus. The frequencies are normalized with 1 corresponding to the most frequent token within each diagnostic condition.}
}
\label{fig7}

\end{figure*}

Next, we conducted an interpretability analysis to understand which tokens the model focuses on when forecasting a specific target disease code. For each test individual, whenever the target code is predicted as the top-1 forecast at any time step, we then identified 15 input tokens having the highest cross-attention with the target code. This process was repeated for every individual in the test cohort, we analyzed the 10 most frequent codes with the highest attention values for a given diagnostic task.
The results for 4 out of 12 diagnostic tasks are shown in FIG.\ref{fig7}. The majority of the highly attended codes are related to diagnoses, but they also include medications, treatments, and demographics such as age. For example, the key codes for predicting heart failure were the diagnosis of atrial fibrillation and flutter, the medication used to treat high blood pressure (hypertension), heart failure, a build-up of fluid in the body (edema), and heart rate measurement. In the prediction of major depressive disorder, the model assigned high attention to diagnostic codes related to obesity, allergic rhinitis, and sleep apnea.

\subsection*{\textbf{Fine-tuning evaluation results}}

Table \ref{tab2} reports the number of true positives (TP) achieved via fine-tuning (TP${_f}$) and in the zero-shot setting (TP$_{z}$) across 12 diagnostic tasks, evaluated at a fixed number of false positives (FP). Overall, while fine-tuning provides performance gains, the absolute differences in TP between TP$_{f}$ and TP$_{z}$ are generally modest. For instance, in the 3-month prediction window, the average TP improvement from fine-tuning across tasks is 3.3, with particularly noticeable gains for diagnoses like I25 (chronic ischemic heart disease: 19.62 vs. 11.84) and J44 (chronic obstructive pulmonary disease: 7.01 vs. 2.67). However, in several other tasks—such as M32 (systemic lupus erythematosus) and C22 (liver cancer)—zero-shot performance is within 0.5–1.0 TP of the fine-tuned counterpart.
Similar trends are observed at the 6-month horizon. Fine-tuning yields higher TPs in most tasks, but the zero-shot model often remains competitive. Notably, in tasks with low prevalence or more complex diagnostic criteria (e.g., C25, pancreatic cancer), both settings achieve low TPs, indicating shared difficulty rather than fine-tuning efficacy. Overall, these results suggest the robustness of the zero-shot representation.
Furthermore, these performance deltas must be interpreted in light of the resource demands associated with extensive hyperparameter tuning for task-specific fine-tuning. In contrast, the zero-shot model operates without task-specific training, making it highly scalable across new tasks.

\begin{table*}[]
\begin{tabular}{lcccccccccccc}
\multicolumn{1}{c}{} & I50  & N18   & E11   & C22  & C71  & C25  & J44  & M06  & M32  & I25   & I63  & F32   \\ \hline
\multicolumn{13}{c}{3 months}                                                                                \\ \hline
TP$_{f}$               & 6.86 & 17.2  & 13.86 & 2.94 & 2.51 & 0.00 & 7.01 & 1.87 & 2.44 & 19.62 & 2.97 & 9.64  \\
TP$_{z}$                 & 4.57 & 9.37  & 10.50 & 3.53 & 2.35 & 0.00 & 2.67 & 2.17 & 2.93 & 11.84 & 0.77 & 5.41  \\
FP                   & 0.23 & 1.19  & 1.39  & 0.00 & 0.00 & 0.00 & 0.28 & 0.07 & 0.06 & 2.03  & 0.12 & 2.01  \\ \hline
\multicolumn{13}{c}{6 months}                                                                                \\ \hline
TP$_{f}$              & 6.86 & 15.05 & 14.4  & 1.03 & 2.47 & 0.00 & 4.87 & 1.29 & 1.45 & 15.12 & 2.42 & 11.18 \\
TP$_{z}$                & 4.57 & 8.40  & 8.32  & 1.47 & 2.13 & 0.00 & 2.67 & 2.15 & 2.04 & 10.03 & 0.77 & 5.41  \\
$FP$                   & 0.23 & 1.44  & 1.63  & 0.00 & 0.00 & 0.00 & 0.36 & 0.11 & 0.09 & 2.34  & 0.18 & 2.10  \\ \hline
\end{tabular}
\caption{TP$_{f}$ denotes the true positives achieved through fine-tuning, while TP$_{z}$ represents the true positives obtained in the zero-shot setting, both evaluated at a fixed number of false positives across 12 different diagnostic tasks.}
\label{tab2}
\end{table*}

\begin{figure*}[ht!]
\includegraphics{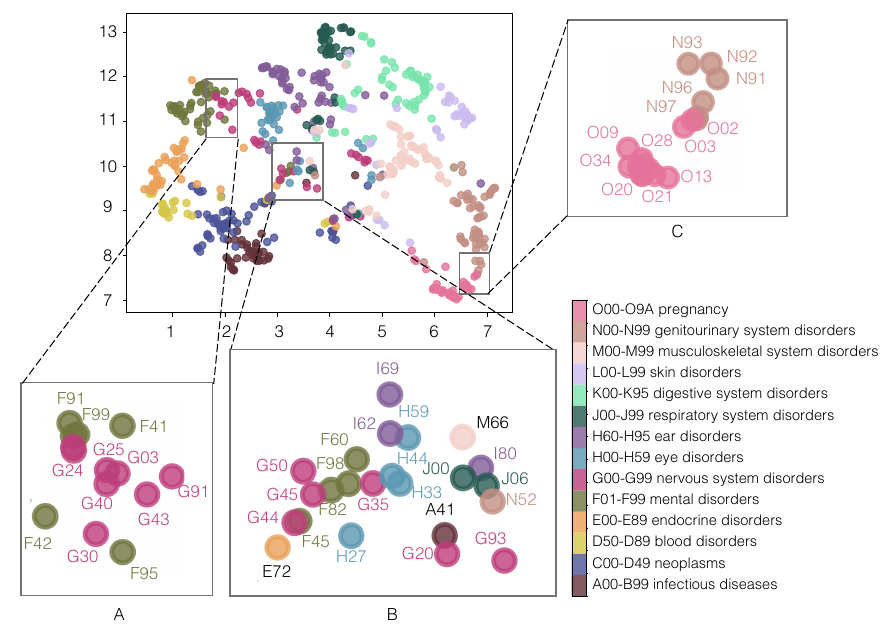}
\caption{\textbf{UMAP of features learned by the GPT model for each code from 14 higher-level categories. A. Example of codes corresponding to mental disorders clustered close to the nervous system disorder codes in the low-dimensional space. B. Example of codes corresponding to various categories clustered together in the low-dimensional space. C. Example of codes corresponding to pregnancy disorders clustered close to the genitourinary system disorders codes in the low-dimensional space.}
}
\label{fig8}
\end{figure*}

\section*{\textbf{Discussion}} 

The model demonstrated promising performance in supporting liver cancer screening efforts, particularly in identifying TP and maintaining a low number of FP, which is critical for this difficult-to-detect disease \citep{yildirim2023advances}. It achieved 46.47\% TP in a 3-month window and 41.18\% in 6 months for the top 20 predictions, indicating its ability to identify cases early and potentially lead to improved treatment outcomes and survival rates \citep{mcmahon2023opportunities}. FP rates remained low at 1. 86\% and 2. 06\% for the 3 and 6-month periods, which could reduce unnecessary interventions and burden. Given the subtle early signs of liver cancer, the performance of the model suggests its potential value in clinical practice \citep{calderaro2022artificial}. Early detection with the model could lead to more timely interventions, improving outcomes.

Early detection of rare diseases such as systemic lupus erythematosus (SLE) is challenging due to the heterogeneity of clinical signs and symptoms \citep{choi2022understanding}. Delayed diagnosis of lupus leads to worse outcomes, including organ damage, increased disease activity, and a lower quality of life \citep{kernder2021delayed}. A model that helps with timely detection is critical to improving outcomes, as early interventions can reduce long-term complications.
Our model demonstrated high performance in zero-shot detection of SLE with 32.07\% TP at $N=20$ within a 3-month window and 31.95\% within 6 months, identifying nearly one-third of the cases. The FP rate was 5.95\% in 3 months and 6.47\% in 6 months, indicating effective identification of true cases while minimizing incorrect diagnoses. A low FP rate is vital to avoid unnecessary tests and treatments and reduce anxiety. The slight increase in FP in the 6-month window suggests extended prediction periods may introduce more uncertainty.

The model's results for forecasting major depressive disorder reveal interesting dynamics that highlight the challenges of detecting depression and the inherent differences from conditions like liver cancer or SLE, where diagnoses tend to be more definitive. For the depression cohort, the model predicted that 6.02\% of individuals would develop depression within the three-month window at $N = 1$. This percentage significantly increased to 84.1\% as $N$ was raised to 20. However, this improvement in TP came at the cost of a substantial increase in FP up to 68.47\%.
 These results indicate a challenge in forecasting depression due to ambiguity in diagnosis, which typically relies on reported symptoms that can vary in intensity and may be underreported or misunderstood\cite{smith2013diagnosis}. Many individuals in the negative cohort, whom the model predicted as FP, could be experiencing undiagnosed depression. The nature of depression, with its wide spectrum of symptoms and the potential for symptoms to change over time, makes it inherently more difficult to classify compared to diseases like liver cancer, where diagnostic tests can often confirm the presence of the disease with greater certainty. Depression can fluctuate between periods of remission and relapse, and individuals may not always present with the same symptoms during medical evaluations \citep{de2019empirical}. This variability could contribute to the model's high FP percentage, especially when $N$ increased, suggesting that the model was capturing patterns indicative of depression but was unable to precisely distinguish between those who would receive a clinical diagnosis and those who may have had undiagnosed depression. 

We used our GPT model to extract learned embeddings for each code across the 14 higher-level code categories. To visualize the relationships within the data, we applied the Uniform Manifold Approximation and Projection (UMAP) technique for dimensionality reduction \cite{mcinnes2018umap}. This allowed us to observe how different code categories clustered together in the low-dimensional space, indicating shared characteristics or relationships within the data (see FIG.\ref{fig8}). We observed that codes within the same group tended to cluster together, with some overlap between different groups. For example, codes related to pregnancy, childbirth, and puerperium ($O00-O9A$) were clustered near codes for diseases of the genitourinary system ($N00-N99$), which reflects a logical connection between these medical categories as the genitourinary system plays a central role in reproductive processes (see FIG.\ref{fig8} A). Conditions or treatments related to pregnancy and childbirth often overlap with or affect the genitourinary system \citep{uzelpasaci2021trimester}. Specifically, conditions such as spontaneous abortion ($O03$) and recurrent pregnancy loss ($N96$) share clinical relevance, as both involve fetal loss and can indicate underlying fertility disorders. Similarly, abnormal uterine bleeding ($N93$) is highly relevant to hemorrhage in early pregnancy ($O20$), given that bleeding disorders often complicate pregnancy \citep{gernsheimer2016congenital}. 

Codes related to mental health disorders and disorders of the nervous system tend to be clustered together due to the strong connection between neurological and psychiatric conditions (see FIG.\ref{fig8} C). Many mental health disorders, such as depression, anxiety, and schizophrenia, are influenced by changes in brain function and neural pathways, which fall under the domain of the nervous system \cite{martin2009neurobiology}. For example, conditions such as Alzheimer's disease ($G30$) are primarily neurological but often have mental health components, such as anxiety ($F42$) \cite{li2014behavioral}. The proximity of these codes in the clustering analysis highlights the shared biological mechanisms and symptoms between mental and neurological disorders, as well as their frequent co-occurrence in individuals.

In the UMAP visualization, some codes are not clustered together with their category but are instead positioned in the middle, creating a mixed distribution (see FIG.\ref{fig8} B). This suggests that these codes represent concepts or categories that share characteristics with multiple groups. For instance, multiple sclerosis ($G35$), a chronic autoimmune disorder that impacts the central nervous system, is defined by a diverse set of symptoms, such as vision impairment, numbness and tingling, localized weakness, bladder and bowel dysfunction, and cognitive impairment \citep{balcer2015vision, lin2019frequency, langdon2011cognition}. Thus, the latent-space position for multiple sclerosis in the UMAP visualization reflects its interconnections with a diverse array of related disorders (e.g. eye disorders, mental disorders, genitourinary disorders). 

Another example is Parkinson's disease ($G20$), a progressive disorder of the nervous system primarily affecting movement control. It is characterized by symptoms such as tremors, bradykinesia, rigidity, and postural instability \citep{tahmasian2017resting, gibb1989significance}. In the UMAP visualization, the ICD code for Parkinson's disease is situated in the middle, reflecting its interconnections with a range of other movement disorders and neurodegenerative conditions. 

Fine-tuning increased the percentage of correctly identified cases for certain diagnostic conditions, in other cases, the zero-shot model already performed competitively. This suggests that while fine-tuning can provide marginal gains, its benefits are not uniform across tasks and may depend on the nature and prevalence of the target condition. Importantly, our fine-tuning setup did not involve full hyperparameter optimization; instead, we used a fixed training configuration to ensure a consistent and practical approach. While performance could likely be further improved with tuning, in our experience, such improvements would be modest and not change the overall interpretation of our results. Moreover, extensive tuning would increase the computational cost.

\subsection*{\textbf{Limitations}} 
Many EHR codes are highly correlated due to the interconnected nature of medical conditions and treatments. Selecting a single code for fine-tuning or zero-shot evaluation can bias results and overestimate model performance. The model may rely on correlations with related codes in an individual’s history rather than demonstrating true causal understanding. For instance, major depressive disorder codes (F32, F33, F34) often follow anxiety disorders (F41), mood disorders (F39), or personality disorders (F60). Similarly, heart failure (I50) is frequently preceded by hypertensive heart disease (I11, I13) or cardiomyopathy (I42).
To better understand these patterns, we performed an error analysis, with detailed results provided in the Supplementary Materials (Figure \ref{umap2}).

When a specific code is not the top prediction, correlated codes often appear among the top results. Thus, a comprehensive evaluation considering multiple codes and their interrelationships is essential for accurately assessing the predictive capabilities of autoregressive models. Finally, ICD codes may not always accurately reflect an individual’s true clinical state, but rather a probabilistic interpretation of the diagnosis based on the information available at the time of coding \cite{o2005measuring}.
Autoregressive models like GPT rely on sequentially ordered data due to their unidirectional attention mechanism. They generate predictions token by token, using prior tokens to guide the output. While EHR data is naturally ordered by visit timestamps, codes within a visit often share identical timestamps, complicating strict sequential ordering. Assigning diagnostic, procedure, medication, and laboratory codes within one visit imposes an artificial order, limiting the model's ability to learn inter-code relationships. Future work will focus on solutions that better capture these relationships within visits.

\section*{\textbf{Conclusion}}
In this work, we developed a GPT model for forecasting EHR concepts using a zero-shot learning approach. This approach offers benefits for healthcare systems by eliminating the need for task-specific fine-tuning, the scaling of which is limited by computational costs, cohort labeling challenges, and ongoing model maintenance.
By predicting various disease occurrences, the model supports a preventive health approach to medicine, allowing hospitals and clinics to intervene prior to the emergence of serious health conditions and thus improve outcomes. The nature of such interventions has become a critical area of study given the improving performance of general models, such as the one presented in this study. Significant future work is needed to understand how to optimize interventions in resource-constrained healthcare environments.

\section*{Contributions}
ER led the conceptualization and methodology with help from ZW, RK, MP, HRH, BH and ME. Supervision of the project was done by CWA, WS, BH, ME and HRH. Data curation was done by ER with help from ZW, AC, and RK. CWA led the project administration and was responsible for the resources. The original draft was written by ER and underwent a thorough review and editing process by ZW, MP, WS, and CWA, each of whom provided invaluable feedback and insights. ZW and RK have accessed and verified all data used for this study. All authors had full access to all the data in the study and had final responsibility for the decision to submit for publication.

\section*{Competing interests}
The authors declare no competing interests.

\bibliographystyle{apsrev4-1}
\bibliography{main}

\begin{thebibliography}{26}%
\makeatletter
\providecommand \@ifxundefined [1]{%
 \@ifx{#1\undefined}
}%
\providecommand \@ifnum [1]{%
 \ifnum #1\expandafter \@firstoftwo
 \else \expandafter \@secondoftwo
 \fi
}%
\providecommand \@ifx [1]{%
 \ifx #1\expandafter \@firstoftwo
 \else \expandafter \@secondoftwo
 \fi
}%
\providecommand \natexlab [1]{#1}%
\providecommand \enquote  [1]{``#1''}%
\providecommand \bibnamefont  [1]{#1}%
\providecommand \bibfnamefont [1]{#1}%
\providecommand \citenamefont [1]{#1}%
\providecommand \href@noop [0]{\@secondoftwo}%
\providecommand \href [0]{\begingroup \@sanitize@url \@href}%
\providecommand \@href[1]{\@@startlink{#1}\@@href}%
\providecommand \@@href[1]{\endgroup#1\@@endlink}%
\providecommand \@sanitize@url [0]{\catcode `\\12\catcode `\$12\catcode `\&12\catcode `\#12\catcode `\^12\catcode `\_12\catcode `\%12\relax}%
\providecommand \@@startlink[1]{}%
\providecommand \@@endlink[0]{}%
\providecommand \url  [0]{\begingroup\@sanitize@url \@url }%
\providecommand \@url [1]{\endgroup\@href {#1}{\urlprefix }}%
\providecommand \urlprefix  [0]{URL }%
\providecommand \Eprint [0]{\href }%
\providecommand \doibase [0]{http://dx.doi.org/}%
\providecommand \selectlanguage [0]{\@gobble}%
\providecommand \bibinfo  [0]{\@secondoftwo}%
\providecommand \bibfield  [0]{\@secondoftwo}%
\providecommand \translation [1]{[#1]}%
\providecommand \BibitemOpen [0]{}%
\providecommand \bibitemStop [0]{}%
\providecommand \bibitemNoStop [0]{.\EOS\space}%
\providecommand \EOS [0]{\spacefactor3000\relax}%
\providecommand \BibitemShut  [1]{\csname bibitem#1\endcsname}%
\let\auto@bib@innerbib\@empty
\bibitem [{\citenamefont {Yang}\ \emph {et~al.}(2023)\citenamefont {Yang}, \citenamefont {Mitra}, \citenamefont {Liu}, \citenamefont {Berlowitz},\ and\ \citenamefont {Yu}}]{yang2023transformehr}%
  \BibitemOpen
  \bibfield  {author} {\bibinfo {author} {\bibfnamefont {Z.}~\bibnamefont {Yang}}, \bibinfo {author} {\bibfnamefont {A.}~\bibnamefont {Mitra}}, \bibinfo {author} {\bibfnamefont {W.}~\bibnamefont {Liu}}, \bibinfo {author} {\bibfnamefont {D.}~\bibnamefont {Berlowitz}}, \ and\ \bibinfo {author} {\bibfnamefont {H.}~\bibnamefont {Yu}},\ }\href@noop {} {\bibfield  {journal} {\bibinfo  {journal} {Nature communications}\ }\textbf {\bibinfo {volume} {14}},\ \bibinfo {pages} {7857} (\bibinfo {year} {2023})}\BibitemShut {NoStop}%
\bibitem [{\citenamefont {Kraljevic}\ \emph {et~al.}(2024)\citenamefont {Kraljevic}, \citenamefont {Bean}, \citenamefont {Shek}, \citenamefont {Bendayan}, \citenamefont {Hemingway}, \citenamefont {Yeung}, \citenamefont {Deng}, \citenamefont {Baston}, \citenamefont {Ross}, \citenamefont {Idowu} \emph {et~al.}}]{kraljevic2024foresight}%
  \BibitemOpen
  \bibfield  {author} {\bibinfo {author} {\bibfnamefont {Z.}~\bibnamefont {Kraljevic}}, \bibinfo {author} {\bibfnamefont {D.}~\bibnamefont {Bean}}, \bibinfo {author} {\bibfnamefont {A.}~\bibnamefont {Shek}}, \bibinfo {author} {\bibfnamefont {R.}~\bibnamefont {Bendayan}}, \bibinfo {author} {\bibfnamefont {H.}~\bibnamefont {Hemingway}}, \bibinfo {author} {\bibfnamefont {J.~A.}\ \bibnamefont {Yeung}}, \bibinfo {author} {\bibfnamefont {A.}~\bibnamefont {Deng}}, \bibinfo {author} {\bibfnamefont {A.}~\bibnamefont {Baston}}, \bibinfo {author} {\bibfnamefont {J.}~\bibnamefont {Ross}}, \bibinfo {author} {\bibfnamefont {E.}~\bibnamefont {Idowu}},  \emph {et~al.},\ }\href@noop {} {\bibfield  {journal} {\bibinfo  {journal} {The Lancet Digital Health}\ }\textbf {\bibinfo {volume} {6}},\ \bibinfo {pages} {e281} (\bibinfo {year} {2024})}\BibitemShut {NoStop}%
\bibitem [{\citenamefont {Pang}\ \emph {et~al.}(2024)\citenamefont {Pang}, \citenamefont {Jiang}, \citenamefont {Pavinkurve}, \citenamefont {Kalluri}, \citenamefont {Minto}, \citenamefont {Patterson}, \citenamefont {Zhang}, \citenamefont {Hripcsak}, \citenamefont {Elhadad},\ and\ \citenamefont {Natarajan}}]{pang2024cehr}%
  \BibitemOpen
  \bibfield  {author} {\bibinfo {author} {\bibfnamefont {C.}~\bibnamefont {Pang}}, \bibinfo {author} {\bibfnamefont {X.}~\bibnamefont {Jiang}}, \bibinfo {author} {\bibfnamefont {N.~P.}\ \bibnamefont {Pavinkurve}}, \bibinfo {author} {\bibfnamefont {K.~S.}\ \bibnamefont {Kalluri}}, \bibinfo {author} {\bibfnamefont {E.~L.}\ \bibnamefont {Minto}}, \bibinfo {author} {\bibfnamefont {J.}~\bibnamefont {Patterson}}, \bibinfo {author} {\bibfnamefont {L.}~\bibnamefont {Zhang}}, \bibinfo {author} {\bibfnamefont {G.}~\bibnamefont {Hripcsak}}, \bibinfo {author} {\bibfnamefont {N.}~\bibnamefont {Elhadad}}, \ and\ \bibinfo {author} {\bibfnamefont {K.}~\bibnamefont {Natarajan}},\ }\href@noop {} {\bibfield  {journal} {\bibinfo  {journal} {arXiv preprint arXiv:2402.04400}\ } (\bibinfo {year} {2024})}\BibitemShut {NoStop}%
\bibitem [{\citenamefont {Meng}\ \emph {et~al.}(2021)\citenamefont {Meng}, \citenamefont {Speier}, \citenamefont {Ong},\ and\ \citenamefont {Arnold}}]{meng2021bidirectional}%
  \BibitemOpen
  \bibfield  {author} {\bibinfo {author} {\bibfnamefont {Y.}~\bibnamefont {Meng}}, \bibinfo {author} {\bibfnamefont {W.}~\bibnamefont {Speier}}, \bibinfo {author} {\bibfnamefont {M.~K.}\ \bibnamefont {Ong}}, \ and\ \bibinfo {author} {\bibfnamefont {C.~W.}\ \bibnamefont {Arnold}},\ }\href@noop {} {\bibfield  {journal} {\bibinfo  {journal} {IEEE journal of biomedical and health informatics}\ }\textbf {\bibinfo {volume} {25}},\ \bibinfo {pages} {3121} (\bibinfo {year} {2021})}\BibitemShut {NoStop}%
\bibitem [{\citenamefont {Rasmy}\ \emph {et~al.}(2021)\citenamefont {Rasmy}, \citenamefont {Xiang}, \citenamefont {Xie}, \citenamefont {Tao},\ and\ \citenamefont {Zhi}}]{rasmy2021med}%
  \BibitemOpen
  \bibfield  {author} {\bibinfo {author} {\bibfnamefont {L.}~\bibnamefont {Rasmy}}, \bibinfo {author} {\bibfnamefont {Y.}~\bibnamefont {Xiang}}, \bibinfo {author} {\bibfnamefont {Z.}~\bibnamefont {Xie}}, \bibinfo {author} {\bibfnamefont {C.}~\bibnamefont {Tao}}, \ and\ \bibinfo {author} {\bibfnamefont {D.}~\bibnamefont {Zhi}},\ }\href@noop {} {\bibfield  {journal} {\bibinfo  {journal} {NPJ digital medicine}\ }\textbf {\bibinfo {volume} {4}},\ \bibinfo {pages} {86} (\bibinfo {year} {2021})}\BibitemShut {NoStop}%
\bibitem [{\citenamefont {Radford}\ \emph {et~al.}(2019)\citenamefont {Radford}, \citenamefont {Wu}, \citenamefont {Child}, \citenamefont {Luan}, \citenamefont {Amodei}, \citenamefont {Sutskever} \emph {et~al.}}]{radford2019language}%
  \BibitemOpen
  \bibfield  {author} {\bibinfo {author} {\bibfnamefont {A.}~\bibnamefont {Radford}}, \bibinfo {author} {\bibfnamefont {J.}~\bibnamefont {Wu}}, \bibinfo {author} {\bibfnamefont {R.}~\bibnamefont {Child}}, \bibinfo {author} {\bibfnamefont {D.}~\bibnamefont {Luan}}, \bibinfo {author} {\bibfnamefont {D.}~\bibnamefont {Amodei}}, \bibinfo {author} {\bibfnamefont {I.}~\bibnamefont {Sutskever}},  \emph {et~al.},\ }\href@noop {} {\bibfield  {journal} {\bibinfo  {journal} {OpenAI blog}\ }\textbf {\bibinfo {volume} {1}},\ \bibinfo {pages} {9} (\bibinfo {year} {2019})}\BibitemShut {NoStop}%
\bibitem [{\citenamefont {Choi}\ \emph {et~al.}(2016)\citenamefont {Choi}, \citenamefont {Bahadori}, \citenamefont {Schuetz}, \citenamefont {Stewart},\ and\ \citenamefont {Sun}}]{choi2016doctor}%
  \BibitemOpen
  \bibfield  {author} {\bibinfo {author} {\bibfnamefont {E.}~\bibnamefont {Choi}}, \bibinfo {author} {\bibfnamefont {M.~T.}\ \bibnamefont {Bahadori}}, \bibinfo {author} {\bibfnamefont {A.}~\bibnamefont {Schuetz}}, \bibinfo {author} {\bibfnamefont {W.~F.}\ \bibnamefont {Stewart}}, \ and\ \bibinfo {author} {\bibfnamefont {J.}~\bibnamefont {Sun}},\ }in\ \href@noop {} {\emph {\bibinfo {booktitle} {Machine learning for healthcare conference}}}\ (\bibinfo {organization} {PMLR},\ \bibinfo {year} {2016})\ pp.\ \bibinfo {pages} {301--318}\BibitemShut {NoStop}%
\bibitem [{\citenamefont {Bellamy}\ \emph {et~al.}(2023)\citenamefont {Bellamy}, \citenamefont {Kumar}, \citenamefont {Wang},\ and\ \citenamefont {Beam}}]{bellamy2023labrador}%
  \BibitemOpen
  \bibfield  {author} {\bibinfo {author} {\bibfnamefont {D.~R.}\ \bibnamefont {Bellamy}}, \bibinfo {author} {\bibfnamefont {B.}~\bibnamefont {Kumar}}, \bibinfo {author} {\bibfnamefont {C.}~\bibnamefont {Wang}}, \ and\ \bibinfo {author} {\bibfnamefont {A.}~\bibnamefont {Beam}},\ }\href@noop {} {\bibfield  {journal} {\bibinfo  {journal} {arXiv preprint arXiv:2312.11502}\ } (\bibinfo {year} {2023})}\BibitemShut {NoStop}%
\bibitem [{\citenamefont {Y{\i}ld{\i}r{\i}m}\ \emph {et~al.}(2023)\citenamefont {Y{\i}ld{\i}r{\i}m}, \citenamefont {Kavgaci}, \citenamefont {Chalabiyev},\ and\ \citenamefont {Dizdar}}]{yildirim2023advances}%
  \BibitemOpen
  \bibfield  {author} {\bibinfo {author} {\bibfnamefont {H.~{\c{C}}.}\ \bibnamefont {Y{\i}ld{\i}r{\i}m}}, \bibinfo {author} {\bibfnamefont {G.}~\bibnamefont {Kavgaci}}, \bibinfo {author} {\bibfnamefont {E.}~\bibnamefont {Chalabiyev}}, \ and\ \bibinfo {author} {\bibfnamefont {O.}~\bibnamefont {Dizdar}},\ }\href@noop {} {\bibfield  {journal} {\bibinfo  {journal} {Cancers}\ }\textbf {\bibinfo {volume} {15}},\ \bibinfo {pages} {3880} (\bibinfo {year} {2023})}\BibitemShut {NoStop}%
\bibitem [{\citenamefont {McMahon}\ \emph {et~al.}(2023)\citenamefont {McMahon}, \citenamefont {Cohen}, \citenamefont {Brown~Jr}, \citenamefont {El-Serag}, \citenamefont {Ioannou}, \citenamefont {Lok}, \citenamefont {Roberts}, \citenamefont {Singal},\ and\ \citenamefont {Block}}]{mcmahon2023opportunities}%
  \BibitemOpen
  \bibfield  {author} {\bibinfo {author} {\bibfnamefont {B.}~\bibnamefont {McMahon}}, \bibinfo {author} {\bibfnamefont {C.}~\bibnamefont {Cohen}}, \bibinfo {author} {\bibfnamefont {R.~S.}\ \bibnamefont {Brown~Jr}}, \bibinfo {author} {\bibfnamefont {H.}~\bibnamefont {El-Serag}}, \bibinfo {author} {\bibfnamefont {G.~N.}\ \bibnamefont {Ioannou}}, \bibinfo {author} {\bibfnamefont {A.~S.}\ \bibnamefont {Lok}}, \bibinfo {author} {\bibfnamefont {L.~R.}\ \bibnamefont {Roberts}}, \bibinfo {author} {\bibfnamefont {A.~G.}\ \bibnamefont {Singal}}, \ and\ \bibinfo {author} {\bibfnamefont {T.}~\bibnamefont {Block}},\ }\href@noop {} {\bibfield  {journal} {\bibinfo  {journal} {JNCI Cancer Spectrum}\ }\textbf {\bibinfo {volume} {7}},\ \bibinfo {pages} {pkad034} (\bibinfo {year} {2023})}\BibitemShut {NoStop}%
\bibitem [{\citenamefont {Calderaro}\ \emph {et~al.}(2022)\citenamefont {Calderaro}, \citenamefont {Seraphin}, \citenamefont {Luedde},\ and\ \citenamefont {Simon}}]{calderaro2022artificial}%
  \BibitemOpen
  \bibfield  {author} {\bibinfo {author} {\bibfnamefont {J.}~\bibnamefont {Calderaro}}, \bibinfo {author} {\bibfnamefont {T.~P.}\ \bibnamefont {Seraphin}}, \bibinfo {author} {\bibfnamefont {T.}~\bibnamefont {Luedde}}, \ and\ \bibinfo {author} {\bibfnamefont {T.~G.}\ \bibnamefont {Simon}},\ }\href@noop {} {\bibfield  {journal} {\bibinfo  {journal} {Journal of hepatology}\ }\textbf {\bibinfo {volume} {76}},\ \bibinfo {pages} {1348} (\bibinfo {year} {2022})}\BibitemShut {NoStop}%
\bibitem [{\citenamefont {Choi}\ and\ \citenamefont {Costenbader}(2022)}]{choi2022understanding}%
  \BibitemOpen
  \bibfield  {author} {\bibinfo {author} {\bibfnamefont {M.~Y.}\ \bibnamefont {Choi}}\ and\ \bibinfo {author} {\bibfnamefont {K.~H.}\ \bibnamefont {Costenbader}},\ }\href@noop {} {\bibfield  {journal} {\bibinfo  {journal} {Frontiers in immunology}\ }\textbf {\bibinfo {volume} {13}},\ \bibinfo {pages} {890522} (\bibinfo {year} {2022})}\BibitemShut {NoStop}%
\bibitem [{\citenamefont {Kernder}\ \emph {et~al.}(2021)\citenamefont {Kernder}, \citenamefont {Richter}, \citenamefont {Fischer-Betz}, \citenamefont {Winkler-Rohlfing}, \citenamefont {Brinks}, \citenamefont {Aringer}, \citenamefont {Schneider},\ and\ \citenamefont {Chehab}}]{kernder2021delayed}%
  \BibitemOpen
  \bibfield  {author} {\bibinfo {author} {\bibfnamefont {A.}~\bibnamefont {Kernder}}, \bibinfo {author} {\bibfnamefont {J.~G.}\ \bibnamefont {Richter}}, \bibinfo {author} {\bibfnamefont {R.}~\bibnamefont {Fischer-Betz}}, \bibinfo {author} {\bibfnamefont {B.}~\bibnamefont {Winkler-Rohlfing}}, \bibinfo {author} {\bibfnamefont {R.}~\bibnamefont {Brinks}}, \bibinfo {author} {\bibfnamefont {M.}~\bibnamefont {Aringer}}, \bibinfo {author} {\bibfnamefont {M.}~\bibnamefont {Schneider}}, \ and\ \bibinfo {author} {\bibfnamefont {G.}~\bibnamefont {Chehab}},\ }\href@noop {} {\bibfield  {journal} {\bibinfo  {journal} {Lupus}\ }\textbf {\bibinfo {volume} {30}},\ \bibinfo {pages} {431} (\bibinfo {year} {2021})}\BibitemShut {NoStop}%
\bibitem [{\citenamefont {Smith}\ \emph {et~al.}(2013)\citenamefont {Smith}, \citenamefont {Renshaw},\ and\ \citenamefont {Bilello}}]{smith2013diagnosis}%
  \BibitemOpen
  \bibfield  {author} {\bibinfo {author} {\bibfnamefont {K.~M.}\ \bibnamefont {Smith}}, \bibinfo {author} {\bibfnamefont {P.~F.}\ \bibnamefont {Renshaw}}, \ and\ \bibinfo {author} {\bibfnamefont {J.}~\bibnamefont {Bilello}},\ }\href@noop {} {\bibfield  {journal} {\bibinfo  {journal} {Comprehensive psychiatry}\ }\textbf {\bibinfo {volume} {54}},\ \bibinfo {pages} {1} (\bibinfo {year} {2013})}\BibitemShut {NoStop}%
\bibitem [{\citenamefont {de~Zwart}\ \emph {et~al.}(2019)\citenamefont {de~Zwart}, \citenamefont {Jeronimus},\ and\ \citenamefont {de~Jonge}}]{de2019empirical}%
  \BibitemOpen
  \bibfield  {author} {\bibinfo {author} {\bibfnamefont {P.~L.}\ \bibnamefont {de~Zwart}}, \bibinfo {author} {\bibfnamefont {B.~F.}\ \bibnamefont {Jeronimus}}, \ and\ \bibinfo {author} {\bibfnamefont {P.}~\bibnamefont {de~Jonge}},\ }\href@noop {} {\bibfield  {journal} {\bibinfo  {journal} {Epidemiology and psychiatric sciences}\ }\textbf {\bibinfo {volume} {28}},\ \bibinfo {pages} {544} (\bibinfo {year} {2019})}\BibitemShut {NoStop}%
\bibitem [{\citenamefont {McInnes}\ \emph {et~al.}(2018)\citenamefont {McInnes}, \citenamefont {Healy},\ and\ \citenamefont {Melville}}]{mcinnes2018umap}%
  \BibitemOpen
  \bibfield  {author} {\bibinfo {author} {\bibfnamefont {L.}~\bibnamefont {McInnes}}, \bibinfo {author} {\bibfnamefont {J.}~\bibnamefont {Healy}}, \ and\ \bibinfo {author} {\bibfnamefont {J.}~\bibnamefont {Melville}},\ }\href@noop {} {\bibfield  {journal} {\bibinfo  {journal} {arXiv preprint arXiv:1802.03426}\ } (\bibinfo {year} {2018})}\BibitemShut {NoStop}%
\bibitem [{\citenamefont {Uzelpasaci}\ \emph {et~al.}(2021)\citenamefont {Uzelpasaci}, \citenamefont {{\c{C}}inar}, \citenamefont {Baran}, \citenamefont {G{\"u}r{\c{s}}en}, \citenamefont {Nakip}, \citenamefont {Ozgul}, \citenamefont {Beksac}, \citenamefont {Unal}, \citenamefont {Orgul}, \citenamefont {Beksac} \emph {et~al.}}]{uzelpasaci2021trimester}%
  \BibitemOpen
  \bibfield  {author} {\bibinfo {author} {\bibfnamefont {E.}~\bibnamefont {Uzelpasaci}}, \bibinfo {author} {\bibfnamefont {G.~N.}\ \bibnamefont {{\c{C}}inar}}, \bibinfo {author} {\bibfnamefont {E.}~\bibnamefont {Baran}}, \bibinfo {author} {\bibfnamefont {C.}~\bibnamefont {G{\"u}r{\c{s}}en}}, \bibinfo {author} {\bibfnamefont {G.}~\bibnamefont {Nakip}}, \bibinfo {author} {\bibfnamefont {S.}~\bibnamefont {Ozgul}}, \bibinfo {author} {\bibfnamefont {K.}~\bibnamefont {Beksac}}, \bibinfo {author} {\bibfnamefont {C.}~\bibnamefont {Unal}}, \bibinfo {author} {\bibfnamefont {G.}~\bibnamefont {Orgul}}, \bibinfo {author} {\bibfnamefont {A.~T.}\ \bibnamefont {Beksac}},  \emph {et~al.},\ }\href@noop {} {\bibfield  {journal} {\bibinfo  {journal} {Current Urology}\ }\textbf {\bibinfo {volume} {15}},\ \bibinfo {pages} {167} (\bibinfo {year} {2021})}\BibitemShut {NoStop}%
\bibitem [{\citenamefont {Gernsheimer}(2016)}]{gernsheimer2016congenital}%
  \BibitemOpen
  \bibfield  {author} {\bibinfo {author} {\bibfnamefont {T.~B.}\ \bibnamefont {Gernsheimer}},\ }\href@noop {} {\bibfield  {journal} {\bibinfo  {journal} {Hematology 2014, the American Society of Hematology Education Program Book}\ }\textbf {\bibinfo {volume} {2016}},\ \bibinfo {pages} {232} (\bibinfo {year} {2016})}\BibitemShut {NoStop}%
\bibitem [{\citenamefont {Martin}\ \emph {et~al.}(2009)\citenamefont {Martin}, \citenamefont {Ressler}, \citenamefont {Binder},\ and\ \citenamefont {Nemeroff}}]{martin2009neurobiology}%
  \BibitemOpen
  \bibfield  {author} {\bibinfo {author} {\bibfnamefont {E.~I.}\ \bibnamefont {Martin}}, \bibinfo {author} {\bibfnamefont {K.~J.}\ \bibnamefont {Ressler}}, \bibinfo {author} {\bibfnamefont {E.}~\bibnamefont {Binder}}, \ and\ \bibinfo {author} {\bibfnamefont {C.~B.}\ \bibnamefont {Nemeroff}},\ }\href@noop {} {\bibfield  {journal} {\bibinfo  {journal} {Psychiatric Clinics}\ }\textbf {\bibinfo {volume} {32}},\ \bibinfo {pages} {549} (\bibinfo {year} {2009})}\BibitemShut {NoStop}%
\bibitem [{\citenamefont {Li}\ \emph {et~al.}(2014)\citenamefont {Li}, \citenamefont {Hu}, \citenamefont {Tan}, \citenamefont {Yu},\ and\ \citenamefont {Tan}}]{li2014behavioral}%
  \BibitemOpen
  \bibfield  {author} {\bibinfo {author} {\bibfnamefont {X.-L.}\ \bibnamefont {Li}}, \bibinfo {author} {\bibfnamefont {N.}~\bibnamefont {Hu}}, \bibinfo {author} {\bibfnamefont {M.-S.}\ \bibnamefont {Tan}}, \bibinfo {author} {\bibfnamefont {J.-T.}\ \bibnamefont {Yu}}, \ and\ \bibinfo {author} {\bibfnamefont {L.}~\bibnamefont {Tan}},\ }\href@noop {} {\bibfield  {journal} {\bibinfo  {journal} {BioMed research international}\ }\textbf {\bibinfo {volume} {2014}},\ \bibinfo {pages} {927804} (\bibinfo {year} {2014})}\BibitemShut {NoStop}%
\bibitem [{\citenamefont {Balcer}\ \emph {et~al.}(2015)\citenamefont {Balcer}, \citenamefont {Miller}, \citenamefont {Reingold},\ and\ \citenamefont {Cohen}}]{balcer2015vision}%
  \BibitemOpen
  \bibfield  {author} {\bibinfo {author} {\bibfnamefont {L.~J.}\ \bibnamefont {Balcer}}, \bibinfo {author} {\bibfnamefont {D.~H.}\ \bibnamefont {Miller}}, \bibinfo {author} {\bibfnamefont {S.~C.}\ \bibnamefont {Reingold}}, \ and\ \bibinfo {author} {\bibfnamefont {J.~A.}\ \bibnamefont {Cohen}},\ }\href@noop {} {\bibfield  {journal} {\bibinfo  {journal} {Brain}\ }\textbf {\bibinfo {volume} {138}},\ \bibinfo {pages} {11} (\bibinfo {year} {2015})}\BibitemShut {NoStop}%
\bibitem [{\citenamefont {Lin}\ \emph {et~al.}(2019)\citenamefont {Lin}, \citenamefont {Butler}, \citenamefont {Boswell-Ruys}, \citenamefont {Hoang}, \citenamefont {Jarvis}, \citenamefont {Gandevia},\ and\ \citenamefont {McCaughey}}]{lin2019frequency}%
  \BibitemOpen
  \bibfield  {author} {\bibinfo {author} {\bibfnamefont {S.~D.}\ \bibnamefont {Lin}}, \bibinfo {author} {\bibfnamefont {J.~E.}\ \bibnamefont {Butler}}, \bibinfo {author} {\bibfnamefont {C.~L.}\ \bibnamefont {Boswell-Ruys}}, \bibinfo {author} {\bibfnamefont {P.}~\bibnamefont {Hoang}}, \bibinfo {author} {\bibfnamefont {T.}~\bibnamefont {Jarvis}}, \bibinfo {author} {\bibfnamefont {S.~C.}\ \bibnamefont {Gandevia}}, \ and\ \bibinfo {author} {\bibfnamefont {E.~J.}\ \bibnamefont {McCaughey}},\ }\href@noop {} {\bibfield  {journal} {\bibinfo  {journal} {PLoS One}\ }\textbf {\bibinfo {volume} {14}},\ \bibinfo {pages} {e0222731} (\bibinfo {year} {2019})}\BibitemShut {NoStop}%
\bibitem [{\citenamefont {Langdon}(2011)}]{langdon2011cognition}%
  \BibitemOpen
  \bibfield  {author} {\bibinfo {author} {\bibfnamefont {D.~W.}\ \bibnamefont {Langdon}},\ }\href@noop {} {\bibfield  {journal} {\bibinfo  {journal} {Current opinion in neurology}\ }\textbf {\bibinfo {volume} {24}},\ \bibinfo {pages} {244} (\bibinfo {year} {2011})}\BibitemShut {NoStop}%
\bibitem [{\citenamefont {Tahmasian}\ \emph {et~al.}(2017)\citenamefont {Tahmasian}, \citenamefont {Eickhoff}, \citenamefont {Giehl}, \citenamefont {Schwartz}, \citenamefont {Herz}, \citenamefont {Drzezga}, \citenamefont {van Eimeren}, \citenamefont {Laird}, \citenamefont {Fox}, \citenamefont {Khazaie} \emph {et~al.}}]{tahmasian2017resting}%
  \BibitemOpen
  \bibfield  {author} {\bibinfo {author} {\bibfnamefont {M.}~\bibnamefont {Tahmasian}}, \bibinfo {author} {\bibfnamefont {S.~B.}\ \bibnamefont {Eickhoff}}, \bibinfo {author} {\bibfnamefont {K.}~\bibnamefont {Giehl}}, \bibinfo {author} {\bibfnamefont {F.}~\bibnamefont {Schwartz}}, \bibinfo {author} {\bibfnamefont {D.~M.}\ \bibnamefont {Herz}}, \bibinfo {author} {\bibfnamefont {A.}~\bibnamefont {Drzezga}}, \bibinfo {author} {\bibfnamefont {T.}~\bibnamefont {van Eimeren}}, \bibinfo {author} {\bibfnamefont {A.~R.}\ \bibnamefont {Laird}}, \bibinfo {author} {\bibfnamefont {P.~T.}\ \bibnamefont {Fox}}, \bibinfo {author} {\bibfnamefont {H.}~\bibnamefont {Khazaie}},  \emph {et~al.},\ }\href@noop {} {\bibfield  {journal} {\bibinfo  {journal} {Cortex}\ }\textbf {\bibinfo {volume} {92}},\ \bibinfo {pages} {119} (\bibinfo {year} {2017})}\BibitemShut {NoStop}%
\bibitem [{\citenamefont {Gibb}\ and\ \citenamefont {Lees}(1989)}]{gibb1989significance}%
  \BibitemOpen
  \bibfield  {author} {\bibinfo {author} {\bibfnamefont {W.}~\bibnamefont {Gibb}}\ and\ \bibinfo {author} {\bibfnamefont {A.}~\bibnamefont {Lees}},\ }\href@noop {} {\bibfield  {journal} {\bibinfo  {journal} {Neuropathology and applied neurobiology}\ }\textbf {\bibinfo {volume} {15}},\ \bibinfo {pages} {27} (\bibinfo {year} {1989})}\BibitemShut {NoStop}%
\bibitem [{\citenamefont {O'malley}\ \emph {et~al.}(2005)\citenamefont {O'malley}, \citenamefont {Cook}, \citenamefont {Price}, \citenamefont {Wildes}, \citenamefont {Hurdle},\ and\ \citenamefont {Ashton}}]{o2005measuring}%
  \BibitemOpen
  \bibfield  {author} {\bibinfo {author} {\bibfnamefont {K.~J.}\ \bibnamefont {O'malley}}, \bibinfo {author} {\bibfnamefont {K.~F.}\ \bibnamefont {Cook}}, \bibinfo {author} {\bibfnamefont {M.~D.}\ \bibnamefont {Price}}, \bibinfo {author} {\bibfnamefont {K.~R.}\ \bibnamefont {Wildes}}, \bibinfo {author} {\bibfnamefont {J.~F.}\ \bibnamefont {Hurdle}}, \ and\ \bibinfo {author} {\bibfnamefont {C.~M.}\ \bibnamefont {Ashton}},\ }\href@noop {} {\bibfield  {journal} {\bibinfo  {journal} {Health services research}\ }\textbf {\bibinfo {volume} {40}},\ \bibinfo {pages} {1620} (\bibinfo {year} {2005})}\BibitemShut {NoStop}%
\end{thebibliography}%

\makeatletter
\setcounter{figure}{0}
\renewcommand \thesection{S\@arabic\c@section}
\renewcommand\thetable{S\@arabic\c@table}
\renewcommand \thefigure{S\@arabic\c@figure}
\makeatother
\section*{Supplementary Materials}

\subsection*{Model Architecture and training configuration}

We pretrained a causal language model on longitudinal EHR sequences using a modified GPT-2 architecture implemented with the HuggingFace Transformers library. The configuration and training setup are summarized below.

\paragraph{Model Architecture.} The architecture details are provided in Table~\ref{tab:model_config}.

\begin{table}[h]
\centering
\begin{tabular}{|l|c|}
\hline
\textbf{Parameter} & \textbf{Value} \\
\hline
Architecture & GPT2LMHeadModel \\
Hidden size & 384 \\
Number of layers & 4 \\
Attention heads & 8 \\
Context window & 513 tokens \\
Vocabulary size & 4840 \\
BOS/EOS token ID & 4839 \\
Total parameters & $\sim$56 million \\
\hline
\end{tabular}
\caption{GPT model configuration used for EHR sequence modeling.}
\label{tab:model_config}
\end{table}

\paragraph{Training Setup.} The model was trained on tokenized patient timelines truncated to 512 tokens per sequence. Each input takes the form:
\[
u_i = (\texttt{[CLS]}, V_1, \texttt{[SEP]}, V_2, \texttt{[SEP]}, \ldots, V_L),
\]
where each visit $V_t = (w_{t,1}, w_{t,2}, \ldots, w_{t,m_t})$ includes diagnosis, medication, procedure, and laboratory tokens.

\begin{itemize}[noitemsep, topsep=0pt]
    \item \textbf{Batch size:} 256
    \item \textbf{Optimizer:} AdamW
    \item \textbf{Learning rate:} 3e-4
    \item \textbf{Epsilon:} 1e-8
    \item \textbf{Training epochs:} 2
    \item \textbf{Precision:} Full FP32
\end{itemize}

\paragraph{Loss Function.} The model was trained using standard causal language modeling loss (next-token prediction), with attention masking applied to ignore padded positions.

\paragraph{Reproducibility.} Model checkpoints and training losses were saved per epoch. The data was split at the patient level to ensure no leakage between training, validation, and test sets.

\subsection*{Error analysis}
\begin{figure*}[ht!]
\includegraphics{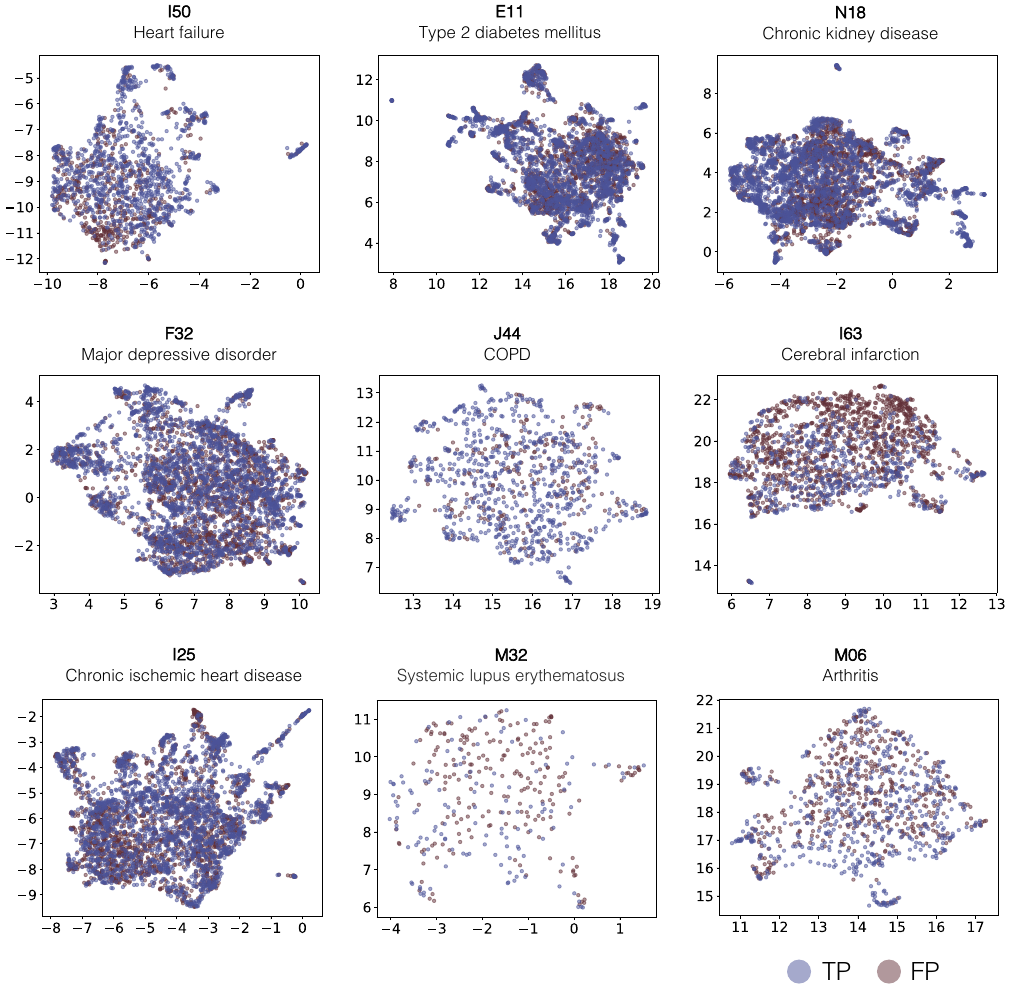}
\caption{\textbf{UMAP visualization of patient embeddings generated by the pretrained GPT model for nine diagnostic conditions. Each point represents an individual patient, colored by prediction outcome (true positive or false positive).}
}
\label{umap2}
\end{figure*}

\begin{figure*}[ht!]
\includegraphics{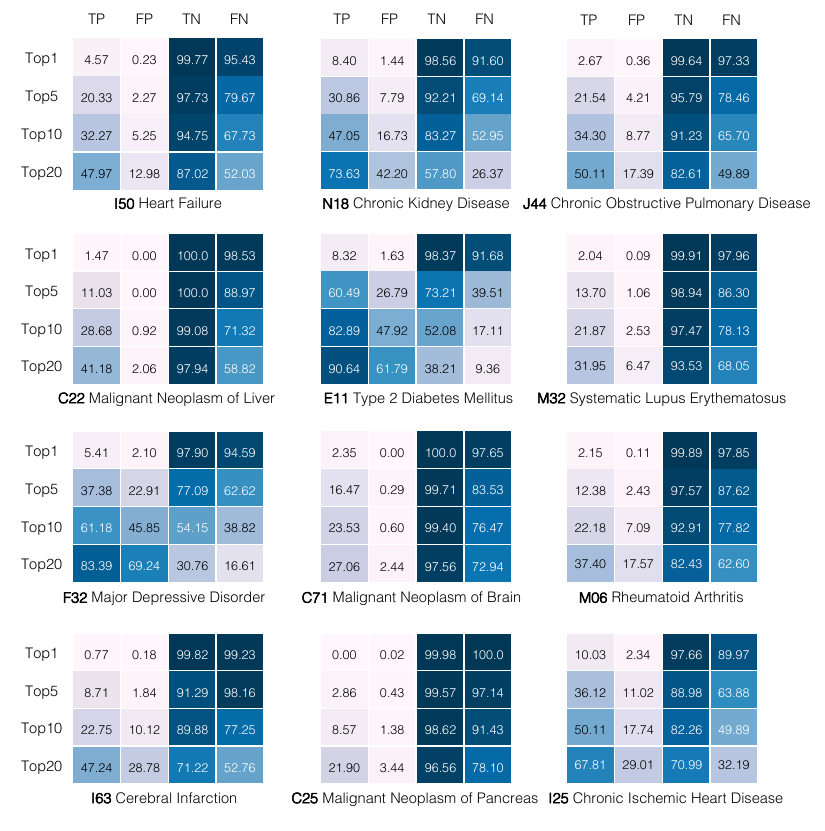}
\caption{\textbf{Zero-shot evaluation results of GPT model for 12 higher-level categories at prediction window of six months. Results are shown using TP,FP,TN and FN in \% for $N=1,5,10,20$.}
}
\label{fig4}
\end{figure*}

\begin{figure*}[ht!]
\includegraphics{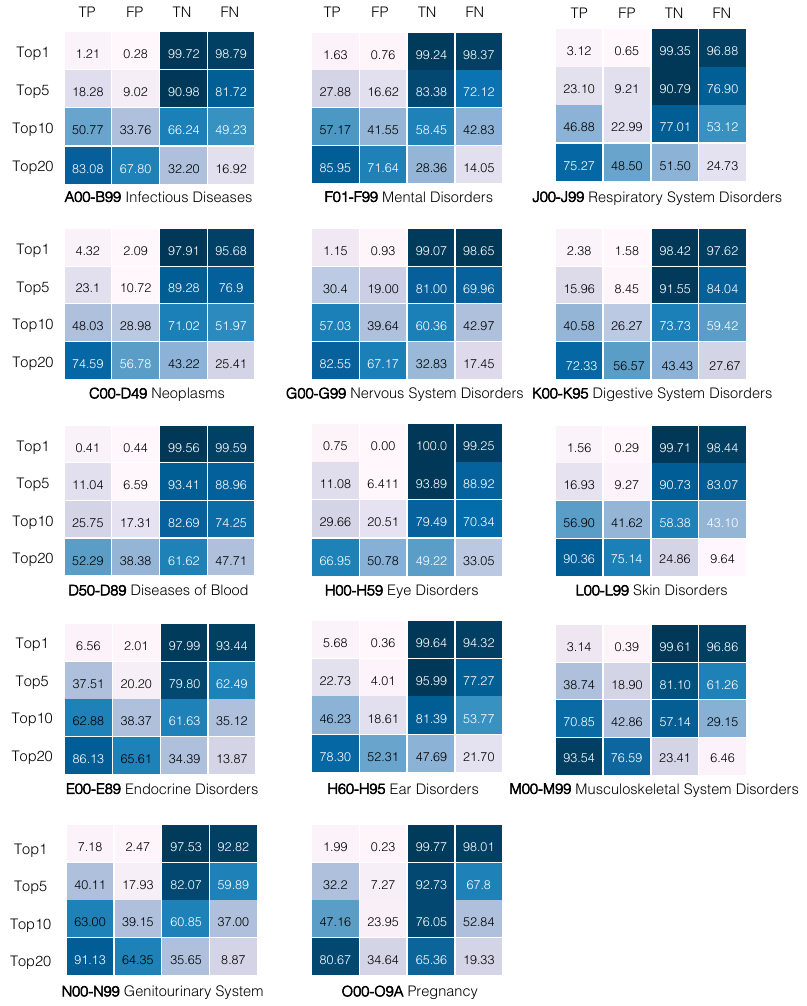}
\caption{\textbf{Zero-shot evaluation results of GPT model for 14 specific diagnostic conditions at prediction window of six months. Results are shown using TP,FP,TN and FN in \% for $N=1,5,10,20$.}
}
\label{fig6}
\end{figure*}

To further explore the behavior of the pretrained GPT model, we visualized patient-level embeddings using UMAP for several diagnostic conditions. For each condition, we extracted the final-layer embeddings generated by the model for individual patients and projected them into a two-dimensional space using UMAP. We then labeled each point as a TP or FP based on the model's zero-shot prediction outcome. As shown in Figure \ref{umap2}, there is substantial overlap between TP and FP patient representations across multiple diseases. This overlap suggests that the model assigns high confidence to patients who are, in fact, clinically similar to true positives, but ultimately do not receive the diagnosis. These errors may stem from limitations in the model's ability to distinguish subtle clinical differences or from ambiguity in the diagnostic labels themselves (e.g., due to underdiagnosis, delayed documentation, or overlapping comorbidities). Future work is needed to better understand these borderline cases and whether incorporating richer supervision signals or patient-level context could reduce such errors.

\label{sec:supplementary}

\
\end{document}